\pdfoutput=1

\documentclass[11pt]{article}

\usepackage{acl} 
\usepackage{bookmark}
\usepackage{times}
\usepackage{latexsym}
\usepackage[T1]{fontenc}
\usepackage[utf8]{inputenc}
\usepackage{microtype}
\usepackage{inconsolata}
\usepackage{graphicx}
\usepackage{tikz}
\usepackage{xcolor}
\usepackage{amsmath}
\usepackage{amssymb}
\usepackage{fontawesome5}
\usepackage{booktabs}
\usepackage{array}
\usepackage{ragged2e}
\usetikzlibrary{shapes,arrows,positioning,shadows}

\definecolor{primaryblue}{HTML}{2563eb}
\definecolor{secondaryblue}{HTML}{1d4ed8}
\definecolor{tertiaryblue}{HTML}{1e40af}
\definecolor{quaternaryblue}{HTML}{1e3a8a}
\definecolor{processorange}{HTML}{f97316}
\definecolor{semanticlight}{HTML}{dcfce7}
\definecolor{semanticmedium}{HTML}{bbf7d0}
\definecolor{semanticstrong}{HTML}{86efac}
\definecolor{semanticdark}{HTML}{16a34a}

\title{Seeing Symbols, Missing Cultures: Probing Vision-Language Models' Reasoning on Fire Imagery and Cultural Meaning}

\newcommand{\authorrowsep}{3.8em} 
\author{%
\begin{minipage}[t]{0.48\textwidth}\centering
{\bfseries\large Haorui Yu}\\[1pt]
{\normalfont\normalsize Duncan of Jordanstone College of\\
Art \& Design (DJCAD), University of Dundee\\
Dundee, United Kingdom}\\[1pt]
{\normalfont\normalsize \texttt{2655435@dundee.ac.uk}}
\end{minipage}\hfill
\begin{minipage}[t]{0.48\textwidth}\centering
{\bfseries\large Yang Zhao}\\[1pt]
{\normalfont\normalsize Guangzhou Institute of Science\\
and Technology (GZIST), Guangzhou, China}\\[1pt]
{\normalfont\normalsize \texttt{zhaoyang@gzist.edu.cn}}
\end{minipage}
\\[\authorrowsep]
\begin{minipage}[t]{0.48\textwidth}\centering
{\bfseries\large Yijia Chu}\\[1pt]
{\normalfont\normalsize Faculty of Arts, Xiamen University\\
Xiamen, China}\\[1pt]
{\normalfont\normalsize \texttt{18620221154827@stu.xmu.edu.cn}}
\end{minipage}\hfill
\begin{minipage}[t]{0.48\textwidth}\centering
{\bfseries\large Qiufeng Yi}\\[1pt]
{\normalfont\normalsize University of Birmingham\\
Birmingham, United Kingdom}\\[1pt]
{\normalfont\normalsize \texttt{qxy953@student.bham.ac.uk}}
\end{minipage}
}

\begin{document}
\maketitle

\begin{abstract}
Vision--Language Models (VLMs) often appear culturally competent but rely on superficial pattern matching rather than genuine cultural understanding. We introduce a \textbf{controlled} diagnostic framework 
to probe VLM reasoning on fire-themed cultural imagery through both classification and explanation analysis. Testing multiple models on Western festivals, non-Western traditions, and emergency scenes reveals systematic biases: models correctly identify prominent Western festivals but struggle with underrepresented cultural events, frequently offering vague labels or misclassifying emergencies as celebrations. 
These failures \textbf{pose risks in public-facing or safety-critical applications} and highlight the need for \textbf{explanation-driven} cultural evaluation beyond accuracy metrics to support interpretable and fair multimodal systems. 
\end{abstract}

\section{Introduction}

Vision-Language Models (VLMs) demonstrate sophisticated capabilities, often appearing culturally aware by correctly identifying festivals and artifacts \cite{sukiennik2025evaluation,liu2025culturevlm}. This apparent competence creates a ``semantic illusion'' where pattern matching masquerades as understanding \cite{li2023evaluating}. A model might label an image as ``Torch Festival'' not from understanding its Yi ethnic significance, but from associating visual cues like fire and crowds with festival tokens. 
This surface-level pattern matching creates critical vulnerabilities \cite{ananthram2024see}.

Common visual elements are semantically ambiguous and culturally polysemous \cite{saussure1916course,turner1967forest}. Fire can signify celebration, crisis, or ritual transformation across cultures \cite{bachelard1964psychoanalysis}. When VLMs use ``symbolic shortcuts''---defaulting to familiar associations rather than contextual specificity---they risk misinterpreting cultural meaning \cite{blodgett2020language}. Models unable to distinguish Peru's sacred Inti Raymi from Britain's Lewes Bonfire, or from dangerous fires, \textbf{pose risks in public-facing or safety-critical applications and therefore warrant additional cultural-robustness evaluation before deployment} \cite{mehrabi2021survey}. 

This paper investigates whether VLMs understand cultural semantics or rely on symbolic shortcuts. We extend recent work on VLM cultural biases \cite{nayak2024culturalvqa,qiu2025cross} with a diagnostic framework probing reasoning patterns. We analyze both classification labels and explanations \cite{ferrara2024fairness}, evaluating models on visually similar but semantically distinct fire-themed images to expose reasoning failures that accuracy metrics miss. Figure~\ref{fig:framework} illustrates our methodology.

Our approach differs from recent frameworks like CROSS \cite{qiu2025cross} in three key ways: (1) we focus on explanation analysis rather than accuracy alone, (2) we \textbf{use ``symbolic shortcuts'' as our diagnostic lens} (rather than claiming a new concept), and (3) we identify safety-critical failure modes when cultural misinterpretation occurs in emergency contexts. 

We formally define \textbf{symbolic shortcuts} as reasoning patterns where models map visual elements (e.g., fire) directly to their most common semantic associations (e.g., festival) while neglecting contextual cues that would enable proper cultural interpretation. \textbf{Rather than a comprehensive benchmark, our contribution is a controlled diagnostic focused on a single multi-meaning symbol (``fire''), complementary to breadth-first cultural evaluations.} 

Our key contributions are: (1) A diagnostic framework that moves beyond accuracy to evaluate VLM reasoning through classification and explanation. (2) A targeted analysis revealing how symbolic shortcuts lead to cultural misinterpretations and safety risks. (3) Evidence of a significant reasoning gap between Western and non-Western cultural contexts, highlighting data bias and fairness issues in state-of-the-art models.

\section{Symbolic Reasoning Probe}

Our diagnostic framework is designed to probe the reasoning behind a VLM's cultural classifications. It assesses whether a model's output is based on genuine semantic understanding or a reliance on superficial visual cues. The probe consists of three components: a curated dataset with controlled semantic ambiguity, a selection of diverse VLMs, and an evaluation protocol that demands both classification and explanation.

\subsection{Model Selection}
We evaluate \textbf{9} recent Vision-Language Models (5 proprietary and 4 open-source), representing a diverse range of architectures and developers. 
This selection allows for a comprehensive comparison across the most capable models available at the time of study.

\begin{table}[t]
\centering
\small
\begin{tabular}{lll}
\toprule
\textbf{Model} & \textbf{Type} & \textbf{Developer} \\
\midrule
GPT-4o & Proprietary & OpenAI \\ 
Claude 3.5 Haiku & Proprietary & Anthropic \\ 
Claude 3.7 Sonnet & Proprietary & Anthropic \\ 
Claude 4 Sonnet & Proprietary & Anthropic \\ 
Claude 4 Opus & Proprietary & Anthropic \\ 
Aya Vision 32B & Open-source & Cohere \\ 
Aya Vision 8B & Open-source & Cohere \\ 
Qwen2.5-VL 72B & Open-source & Alibaba \\ 
Qwen2.5-VL 7B & Open-source & Alibaba \\ 
\bottomrule
\end{tabular}
\caption{Vision--Language Models evaluated (\textbf{9} total; 5 proprietary, 4 open-source).} 
\label{tab:models}
\end{table}

\subsection{Dataset}
To test the models' ability to handle symbolic ambiguity, we curated a Multi-Cultural Heritage Dataset (MCHD) of 77 images. The images are thematically consistent (fire-related) but semantically diverse, organized into three categories designed to challenge superficial visual reasoning:
\begin{itemize}
    \item \textbf{Modern Western Festivals (e.g., Burning Man, Guy Fawkes Night):} Events with extensive documentation and high representation in typical training data.
    \item \textbf{Underrepresented Non-Western Traditions (e.g., Huobajie, Sadeh, Inti Raymi):} Events that are visually similar to Western festivals but have distinct cultural meanings and are less likely to be well-represented in training corpora.
    \item \textbf{Non-Cultural Emergencies (e.g., wildfires, structural fires):} Scenes containing fire and sometimes crowds, serving as a critical control group to test for cultural misattribution and safety-critical failures.
\end{itemize}
A detailed list of the cultural traditions included is available in Appendix~\ref{sec:appendix_categories}.

The 77 images were sourced from publicly available online repositories under Creative Commons licenses. Selection criteria included: (1) clear fire-related visual elements, (2) sufficient contextual cues for cultural identification, (3) resolution suitable for VLM processing (minimum 512$\times$512 pixels), and (4) verification of cultural authenticity through multiple sources. 
The distribution comprises: 30 Western festival images, 37 non-Western tradition images, and 10 emergency control images.

\paragraph{Availability.}
To prevent test contamination and overfitting in future model training, we keep the test images private while releasing metadata (URLs, licenses, cultural labels) and the full evaluation scripts/prompts in the supplementary material. 

\subsection{Evaluation Protocol}
We use a single, zero-shot direct prompt for both classification and explanation: ``Please identify the cultural event or tradition shown in this image. Provide a specific name and general category.'' This simple prompt \textbf{aims to minimize prompt-engineering confounds and tests inherent reasoning without tailored instructions}. 
We acknowledge that our prompt ``Please identify the cultural event or tradition'' may introduce bias by priming models toward cultural interpretations. Future work will explore more neutral prompts such as ``What is shown in this image?'' to reduce presuppositional influences on model responses.

We then manually analyzed the textual responses following these criteria:
\begin{itemize}
\item \textbf{Symbolic shortcuts:} When models rely on generic visual features (e.g., ``fire equals festival'')
\item \textbf{Cultural-specific knowledge:} When explanations include specific cultural details
\end{itemize}
Analysis was conducted independently by two evaluators with inter-rater reliability of 0.87 (Cohen's kappa), with disagreements resolved through discussion.

\paragraph{Design rationale (stress test).}
We intentionally use a presuppositional single-step prompt as a \textit{worst-case stress test} that mirrors common user flows and probes whether models can resist cultural priming when the image is in fact an emergency. Without changing prompts, we also report GPT-4o results for two releases (08--06 and 11--20) alongside their aggregate; the qualitative and quantitative patterns are consistent across versions (Tables~\ref{tab:per_category_accuracy}, \ref{tab:appendix_benchmark_p1}, \ref{tab:appendix_benchmark_p2}). 

Due to the subjective nature of cultural recognition and the specialized knowledge required, establishing human baselines is beyond the scope of this diagnostic study.

\section{Findings}

State-of-the-art VLMs consistently favor symbolic shortcuts over cultural reasoning, evident in qualitative output differences and varied performance across categories.

Table~\ref{tab:qualitative_analysis} contrasts GPT-4o and Qwen2.5-VL 72B outputs. Both correctly identify Burning Man, but for Huobajie, GPT-4o identifies the Yi ethnic tradition while Qwen provides only ``Bonfire festival''---demonstrating cultural knowledge gaps. 

Three primary failure modes emerge: (1) \textbf{Cultural Misclassification}---labeling emergencies as cultural events; (2) \textbf{Generic Labeling}---using vague descriptors; (3) \textbf{Western-centric Bias}---defaulting to familiar Western events.

Critically, Qwen misinterprets a wildfire as Guy Fawkes Night---a safety-critical failure where models hallucinate familiar cultural contexts onto dangerous events (see Appendix~\ref{sec:appendix_misclassification}).

Tables~\ref{tab:per_category_accuracy} and \ref{tab:qualitative_analysis} quantify this imbalance and demonstrate the qualitative differences. Performance on \texttt{burning\_man\_american} reaches 100\%, but drops to 0\% for \texttt{sadeh\_iranian} and \texttt{huobajie}, revealing bias toward Western, internet-prominent events.

\begin{figure*}[t]
\centering
\includegraphics[width=0.42\textwidth]{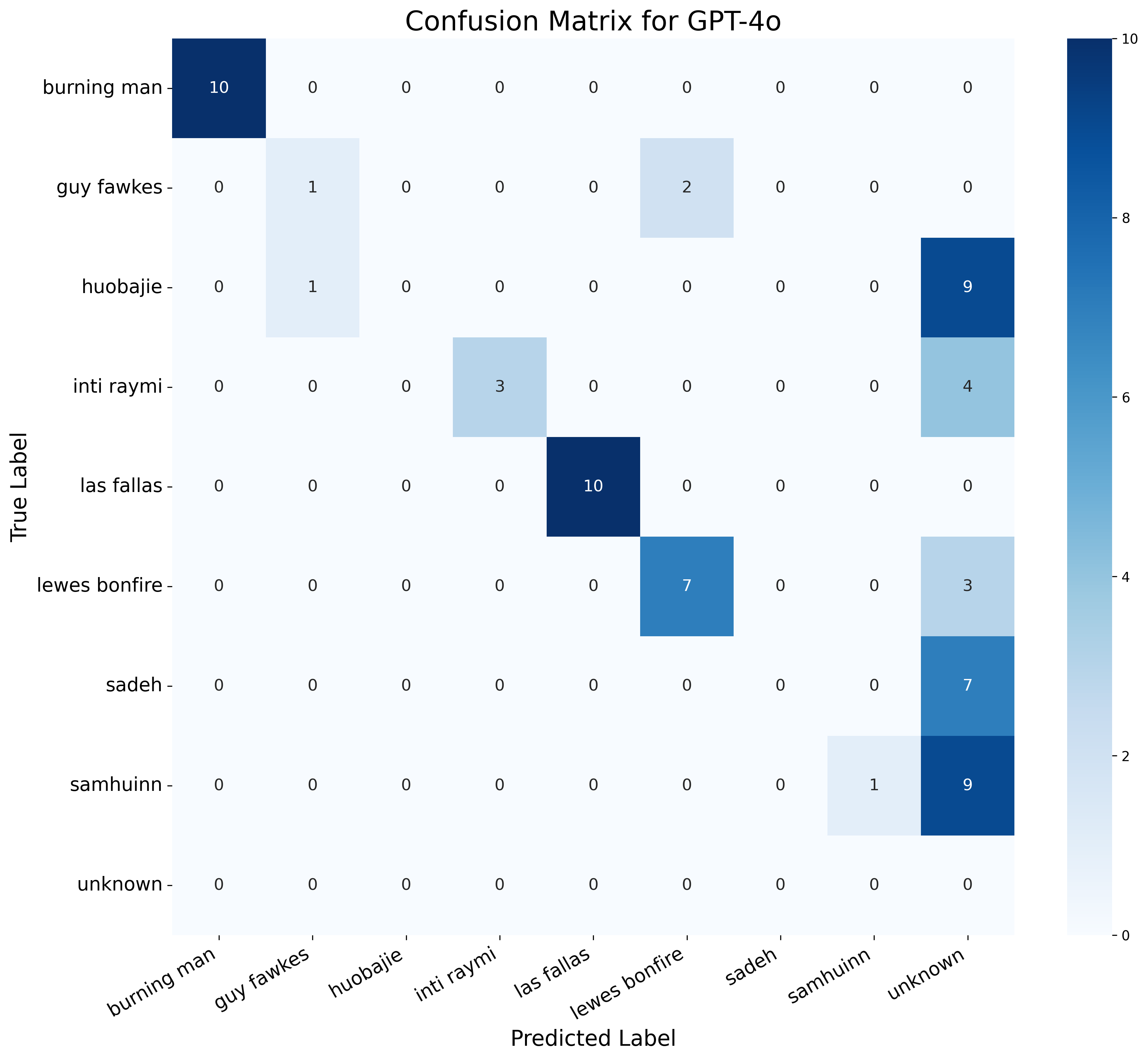}
\includegraphics[width=0.42\textwidth]{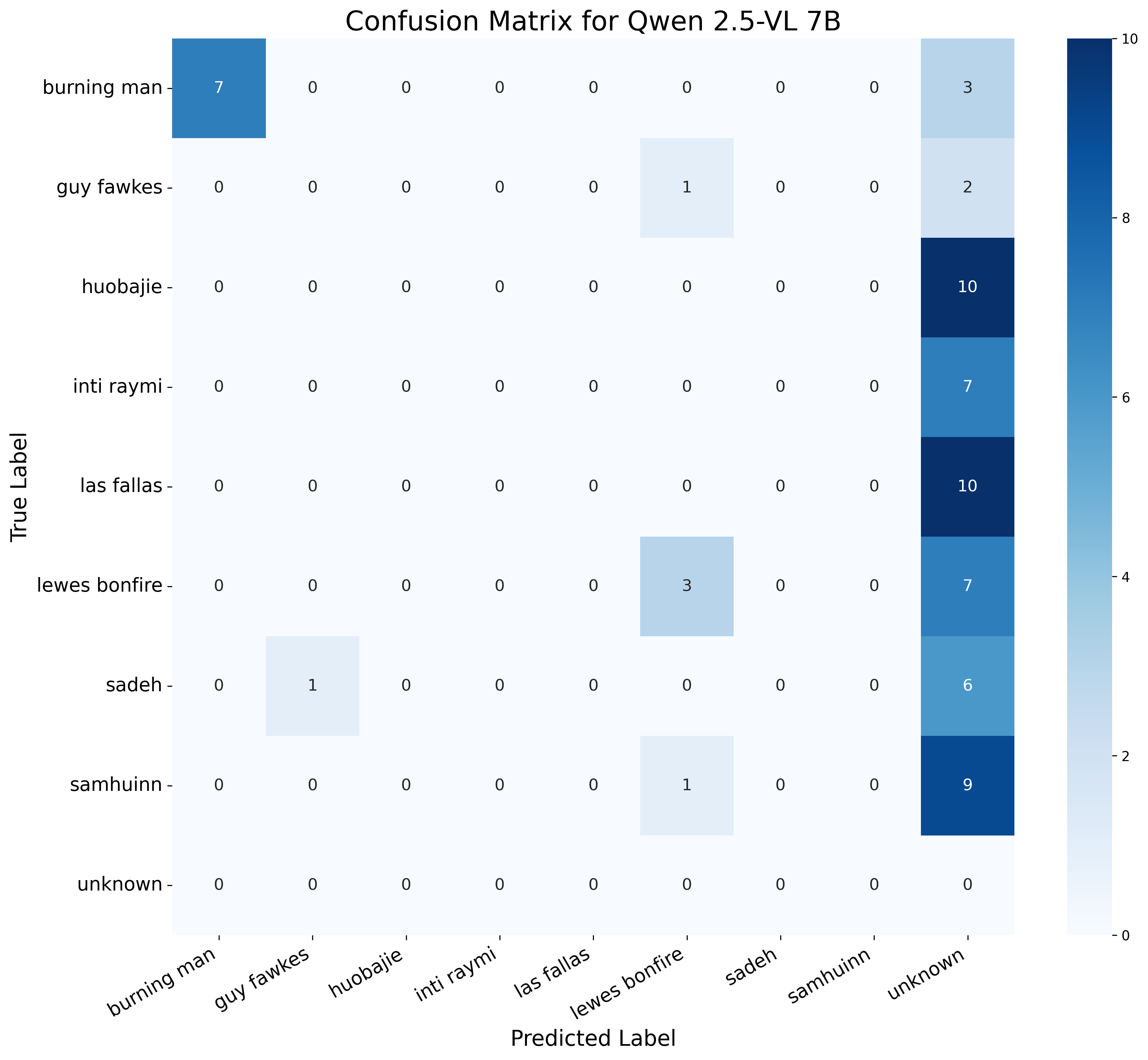}
\caption{Confusion Matrices for GPT-4o (left) and Qwen 2.5-VL 7B (right). The rows represent the true cultural labels, and the columns represent the predicted labels. These matrices reveal the specific patterns of misclassification for the highest and lowest-performing models.}
\label{fig:confusion_matrices}
\end{figure*}

Figure~\ref{fig:confusion_matrices} visualizes failure patterns. GPT-4o shows distributed errors, while Qwen2.5-VL 7B systematically defaults to \texttt{guy fawkes} or \texttt{burning man} when uncertain---\textit{consistent with} reliance on symbolic shortcuts. 

\begin{table*}[!htbp]
\centering
\footnotesize
\renewcommand{\arraystretch}{1.2}
\begin{minipage}[t]{0.36\textwidth}
\centering
\scriptsize
\setlength{\tabcolsep}{4pt}
\begin{tabular}{@{}lccc@{}}
\toprule
\textbf{Tradition} & \textbf{GPT-4o} & \textbf{Claude} & \textbf{Qwen}* \\
\midrule
Burning Man & \textbf{100.0} & 80.0 & 80.0 \\
Guy Fawkes & 33.3 & \textbf{66.7} & 0.0 \\
Huobajie & 0.0 & 0.0 & 0.0 \\
Inti Raymi & \textbf{42.9} & 28.6 & 14.3 \\
Las Fallas & \textbf{100.0} & \textbf{100.0} & 50.0 \\
Lewes Bonfire & \textbf{70.0} & 50.0 & 20.0 \\
Sadeh & 0.0 & \textbf{14.3} & 0.0 \\
Samhuinn & \textbf{10.0} & 0.0 & 10.0 \\
\bottomrule
\end{tabular}
\caption{Fine-grained accuracy by cultural category (\%). Top-tier proprietary (GPT-4o), mid-tier proprietary (Claude 3.7), and open-source (Qwen2.5-VL) models. Best per category in bold. *72B version.}
\label{tab:per_category_accuracy}
\end{minipage}
\hfill
\begin{minipage}[t]{0.62\textwidth}
\centering
\scriptsize
\setlength{\tabcolsep}{3pt}
\begin{tabular}{@{}p{1.3cm}cp{2.5cm}p{2.5cm}@{}}
\hline
\textbf{Case\newline Type} & \textbf{Image} & \textbf{GPT-4o (Top-tier)} & \textbf{Qwen2.5-VL 72B} \\
\hline
\textbf{Western} & 
\raisebox{-0.5\height}{\includegraphics[width=1.8cm]{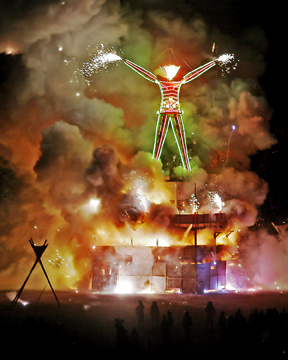}} & 
\begin{tabular}[t]{@{}p{2.4cm}@{}}
\textbf{Name:} Burning Man\\
\textbf{Category:} Art/music\\
\textbf{Analysis:} Correct
\end{tabular} &
\begin{tabular}[t]{@{}p{2.4cm}@{}}
\textbf{Name:} Burning Man\\
\textbf{Category:} Music\\
\textbf{Analysis:} Less specific
\end{tabular} \\
\hline
\textbf{Non-\newline Western} & 
\raisebox{-0.5\height}{\includegraphics[width=1.8cm]{figures/huobajie_chinese/huobajie_chinese_01.jpg}} & 
\begin{tabular}[t]{@{}p{2.4cm}@{}}
\textbf{Name:} Huobajie\\
\textbf{Category:} Yi ethnic\\
\textbf{Analysis:} Accurate
\end{tabular} &
\begin{tabular}[t]{@{}p{2.4cm}@{}}
\textbf{Name:} Bonfire fest.\\
\textbf{Category:} Traditional\\
\textbf{Analysis:} Generic
\end{tabular} \\
\hline
\textbf{Emergency} & 
\raisebox{-0.5\height}{\includegraphics[width=1.8cm]{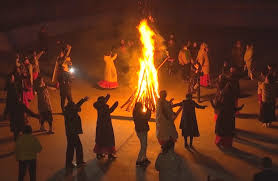}} & 
\begin{tabular}[t]{@{}p{2.4cm}@{}}
\textbf{Name:} Wildfire\\
\textbf{Category:} Emergency\\
\textbf{Analysis:} Correct
\end{tabular} &
\begin{tabular}[t]{@{}p{2.4cm}@{}}
\textbf{Name:} Guy Fawkes\\
\textbf{Category:} Festival\\
\textbf{Analysis:} Dangerous
\end{tabular} \\
\hline
\end{tabular}
\caption{Qualitative case study comparing model outputs on three image types, showing disparities in specificity, cultural knowledge, and safety-critical distinctions. \textbf{Ground Truth:} Western: Burning Man; Non-Western: Yi Torch Festival (Huobajie); Emergency: \textbf{Uncontrolled large-scale outdoor fire (non-cultural)}.} 
\label{tab:qualitative_analysis}
\end{minipage}
\end{table*}

\section{Discussion}

Our findings reveal a gap between visual pattern recognition and cultural understanding. Models' ``symbolic shortcuts''---overgeneralizing fire as festival---create competence illusions masking reasoning failures.

Data imbalance drives these failures. Superior performance on Burning Man versus poor results on Huobajie and Sadeh reveals Western-centric training data bias \cite{ferrara2024fairness}. Models learn simplified dominant representations, not varied cultural meanings.

\paragraph{Mechanistic hypothesis (post hoc).}
The confusion patterns (Fig.~\ref{fig:confusion_matrices}) suggest that, under uncertainty, some models disproportionately map inputs to frequent Western tokens (e.g., \texttt{guy fawkes}, \texttt{burning man}), a ``shortcut prior'' in which co-occurring proxies (flames, crowd density, nighttime) outweigh contextual cues. This is consistent with spurious-correlation phenomena discussed in fairness/bias surveys~\cite{ferrara2024fairness,mehrabi2021survey,blodgett2020language}. A full mechanistic dissection (e.g., attribution analyses) is beyond our diagnostic scope and left for future work. 

This causes cultural erasure---labeling Celtic Samhuinn as ``bonfire''---and safety failures---misclassifying wildfires as festivals. 
Systems unable to distinguish celebration from catastrophe \textbf{pose risks in public-facing or safety-critical applications and therefore warrant additional cultural-robustness evaluation before deployment}. 

We must shift from accuracy to interpretability, probing \textit{why} models conclude. Scaling current approaches reinforces biases; future work needs cultural context modeling and reasoning-focused evaluation.

\subsection{Future Directions}
Future research should explore: (1) extending this framework to other cultural domains (clothing, architecture, cuisine) to validate generalizability, (2) developing training methods to mitigate symbolic shortcuts through culture-aware data augmentation, and (3) integrating cultural knowledge graphs into VLM architectures for enhanced contextual reasoning.

\section{Conclusion}

This paper introduced a diagnostic probe to move beyond accuracy-based evaluation and assess the cultural reasoning of Vision-Language Models. Our findings reveal that current models, including state-of-the-art systems like GPT-4o, often rely on ``symbolic shortcuts,'' leading to a superficial understanding that fails in nuanced, non-Western, or safety-critical contexts. They can see the symbols, but they often miss the culture.

We argue for a crucial transition in how we evaluate AI systems for cultural tasks: a shift from measuring \textit{what} they classify to understanding \textit{how} and \textit{why} they reason. This explanation-driven approach is essential for identifying fairness risks associated with data bias and for building models that are not only accurate but also genuinely and safely culturally aware. This work provides a framework and a baseline for this necessary next step in AI development.

\section*{Limitations}

Our narrow focus on fire festivals ensures consistency but limits generalization to other cultural domains. The 77-image sample, while sufficient to demonstrate our diagnostic framework's validity, constrains the universality of our conclusions. This work should be viewed as a proof-of-concept for a diagnostic tool rather than a comprehensive evaluation of VLM cultural understanding.

The single-prompt evaluation approach, though revealing, presents opportunities for expansion with varied prompting strategies. Future work could explore prompt variations to assess their impact on cultural recognition. Additionally, broader cultural domains (clothing, architecture, cuisine) and larger datasets would strengthen the generalizability of our findings.

\section*{Acknowledgments}

We acknowledge the cultural communities whose traditions form this research foundation and thank cultural consultants for their expertise. We appreciate WiNLP's inclusive research environment.

\bibliography{winlp2025_paper}

\clearpage
\appendix

\section{Appendix: Supplementary Materials}
\label{sec:appendix}

\subsection{Cultural Categories in MCHD}
\label{sec:appendix_categories}
The Multi-Cultural Heritage Dataset (MCHD) used in this study includes images from the following cultural traditions and control category.
\begin{itemize}
    \item \textbf{Western Traditions:} Burning Man (American), Guy Fawkes Night (British), Las Fallas (Spanish), Lewes Bonfire (English), Samhuin (Scottish).
    \item \textbf{Non-Western Traditions:} Huobajie (Chinese), Inti Raymi (Peruvian), Sadeh (Iranian).
    \item \textbf{Control Group:} Fire Emergencies (e.g., wildfires, structural fires).
\end{itemize}

\subsection{Qualitative Comparison of Model Explanations}
\label{sec:appendix_qualitative}
Table~\ref{tab:appendix_qualitative} provides representative model explanations demonstrating the contrast between symbolic shortcuts and cultural understanding.
\begin{table*}[htbp]
\centering
\small
\begin{tabular}{p{2.8cm}|p{2.8cm}|p{7.9cm}}
\toprule
\textbf{Image Type} & \textbf{Model} & \textbf{Full Model Explanation (Illustrative)} \\
\midrule
\textbf{Non-Western Tradition} \par (Huobajie) & GPT-4o & \textbf{Prediction:} Huobajie (Torch Festival), Folk festival of the Yi people. \par \textbf{Reasoning:} The image displays elements consistent with the Torch Festival, including large bonfires, traditional clothing worn by participants that resembles Yi ethnic attire, and a celebratory nighttime atmosphere unique to this cultural event. \\
\cmidrule{2-3}
 & Qwen2.5-VL 72B & \textbf{Prediction:} Bonfire festival, Traditional festival. \par \textbf{Reasoning:} This image shows a large bonfire at night with many people gathered around. These are typical features of a bonfire festival. \\
\midrule
\textbf{Non-Cultural Emergency} \par (Wildfire) & GPT-4o & \textbf{Prediction:} Forest fire / Wildfire, Emergency event. \par \textbf{Reasoning:} The image depicts an uncontrolled fire spreading through a forest. This is a characteristic scene of a wildfire, which is a natural disaster, not a cultural event. \\
\cmidrule{2-3}
 & Qwen2.5-VL 72B & \textbf{Prediction:} Guy Fawkes Night, Festival. \par \textbf{Reasoning:} The large fire in the image is reminiscent of the bonfires traditionally lit during Guy Fawkes Night celebrations in the UK. The event appears to be a public gathering. \\
\bottomrule
\end{tabular}
\caption{Qualitative comparison of full textual explanations generated by a top-tier proprietary model (GPT-4o) and a leading open-source model (Qwen2.5-VL 72B). The examples illustrate GPT-4o's ability to cite specific cultural knowledge versus Qwen's reliance on generic visual cues, which leads to critical misclassification of an emergency.}
\label{tab:appendix_qualitative}
\end{table*}

\subsection{Systematic Cultural Misclassification of Emergencies}
\label{sec:appendix_misclassification}
Critical safety failures occurred when models misinterpreted emergency scenes as cultural events. The following table documents instances where VLMs classified fire emergencies as festivals, highlighting the danger of symbolic shortcuts in safety-critical applications.
\begin{table*}[htbp]
\centering
\begin{tabular}{p{3cm}p{2cm}p{2.5cm}p{4.5cm}}
\toprule
\textbf{Model} & \textbf{Error Type} & \textbf{Prediction} & \textbf{Impact} \\
\midrule
\multicolumn{4}{l}{\textit{Cultural Misclassification (Safety-Critical)}} \\
Claude 3.7 Sonnet & Emergency & Las Fallas & Misinterprets danger as celebration \\
Aya Vision 8B & Emergency & Guy Fawkes & Could delay emergency response \\
Aya Vision 32B & Emergency & Guy Fawkes & Could delay emergency response \\
Qwen2.5-VL 7B & Wildfire & Guy Fawkes & Misses critical safety context \\
Qwen2.5-VL 72B & Wildfire & Burning Man & Normalizes dangerous situation \\
\bottomrule
\end{tabular}
\caption{Instances of Cultural Misclassification where models incorrectly identified non-cultural fire emergencies as cultural festivals. This table highlights the safety-critical implications of these failures, which are particularly prevalent in open-source models.}
\label{tab:appendix_misclassification}
\end{table*}

\subsection{Comprehensive Fine-Grained Accuracy Benchmark}
\label{sec:appendix_full_results}
Full performance metrics across all \textbf{9} and 8 cultural traditions plus control group are presented below. These results demonstrate systematic biases toward Western, internet-prominent events.

\begin{table*}[htbp]
\centering
\begin{tabular}{l|cccc|c}
\toprule
\textbf{Proprietary Models} & \textbf{Burning Man} & \textbf{Guy Fawkes} & \textbf{Huobajie} & \textbf{Inti Raymi} & \textbf{Avg.} \\
\midrule
GPT-4o* & 100.0 & 33.3 & 0.0 & 42.9 & 44.1 \\
Claude 3.7 Sonnet & 80.0 & 66.7 & 0.0 & 28.6 & 43.8 \\
Claude 4 Opus & 100.0 & 66.7 & 0.0 & 28.6 & 48.8 \\
Claude 3.5 Haiku & 80.0 & 66.7 & 0.0 & 28.6 & 43.8 \\
Claude 4 Sonnet & 100.0 & 100.0 & 0.0 & 14.3 & 53.6 \\
\midrule
\textbf{Open-Source Models} & & & & & \\
\midrule
Aya Vision 32B & 60.0 & 0.0 & 0.0 & 0.0 & 15.0 \\
Aya Vision 8B & 80.0 & 33.3 & 0.0 & 28.6 & 35.5 \\
Qwen2.5-VL 72B & 80.0 & 0.0 & 0.0 & 14.3 & 23.6 \\
Qwen2.5-VL 7B & 80.0 & 0.0 & 0.0 & 0.0 & 20.0 \\
\bottomrule
\end{tabular}
\caption{Performance comparison on cultural categories (Part 1). *GPT-4o results averaged across versions.}
\label{tab:appendix_benchmark_p1}
\end{table*}

\begin{table*}[htbp]
\centering
\begin{tabular}{l|cccc|c}
\toprule
\textbf{Model} & \textbf{Las Fallas} & \textbf{Lewes Bonfire} & \textbf{Sadeh} & \textbf{Samhuinn} & \textbf{Emergencies} \\
\midrule
\textbf{Proprietary} & & & & & \\
GPT-4o* & 100.0 & 70.0 & 4.8 & 6.7 & 100.0 \\
Claude 3.7 Sonnet & 100.0 & 50.0 & 14.3 & 0.0 & 90.0 \\
Claude 4 Opus & 100.0 & 40.0 & 0.0 & 0.0 & 100.0 \\
Claude 3.5 Haiku & 100.0 & 40.0 & 0.0 & 0.0 & 100.0 \\
Claude 4 Sonnet & 100.0 & 20.0 & 0.0 & 0.0 & 90.0 \\
\midrule
\textbf{Open-Source} & & & & & \\
Aya Vision 32B & 25.0 & 10.0 & 0.0 & 0.0 & 90.0 \\
Aya Vision 8B & 0.0 & 0.0 & 0.0 & 0.0 & 80.0 \\
Qwen2.5-VL 72B & 50.0 & 20.0 & 0.0 & 10.0 & 70.0 \\
Qwen2.5-VL 7B & 0.0 & 30.0 & 0.0 & 0.0 & 80.0 \\
\bottomrule
\end{tabular}
\caption{Performance comparison on cultural categories (Part 2) and emergency control set.}
\label{tab:appendix_benchmark_p2}
\end{table*}

\end{document}